\documentclass[11pt]{article}
\usepackage{acl2016}
\usepackage{times}
\usepackage{latexsym}

\usepackage{graphicx}
\usepackage{lingmacros}

\aclfinalcopy 

\usepackage[bordercolor=white,backgroundcolor=green!20,linecolor=black,colorinlistoftodos]{todonotes}

\title{Neural Network Models for Implicit Discourse Relation Classification in English and Chinese without Surface Features}

\author{Attapol T. Rutherford\thanks{\ \ Work performed while being a student at Brandeis}\\
Yelp\\ 
San Francisco, CA, USA\\
{\tt teruth@yelp.com} \\
\And
Vera Demberg \\
University of Saarland \\
Saarbrucken, Germany\\
{\tt vera@coli.uni-saarland.de}
\And
Nianwen Xue\\ 
Brandeis University\\ 
Waltham, MA, USA\\
{\tt xuen@brandeis.edu}}

\date{}

\begin{document}

\maketitle

\begin{abstract}
Inferring implicit discourse relations in natural language text is the most difficult subtask in discourse parsing. Surface features achieve good performance, but they are not readily applicable to other languages without semantic lexicons. Previous neural models require parses, surface features, or a small label set to work well. Here, we propose neural network models that are based on feedforward and long-short term memory architecture without any surface features. To our surprise, our best configured feedforward architecture outperforms LSTM-based model in most cases despite thorough tuning. Under various fine-grained label sets and a cross-linguistic setting, our feedforward models perform consistently better or at least just as well as systems that require hand-crafted surface features. Our models present the first neural Chinese discourse parser in the style of Chinese Discourse Treebank, showing that our results hold cross-linguistically. 
\end{abstract}

\section{Introduction}
The discourse structure of a natural language text has been analyzed and conceptualized under various frameworks \cite{mann1988rhetorical,lascarides2007segmented,Prasad:2008ww}. The Penn Discourse TreeBank (PDTB) and the Chinese Discourse Treebank (CDTB), currently the largest corpora annotated with discourse structures in English and Chinese respectively, view the discourse structure of a text as a set of discourse relations \cite{Prasad:2008ww,cdtb}. Each discourse relation is grounded by a discourse connective taking two text segments as arguments \cite{Prasad:2008ww}. Implicit discourse relations are those where discourse connectives are omitted from the text and yet the discourse relations still hold. 

While classifying explicit discourse relations is relatively easy, as the discourse connective itself provides a strong cue for the discourse relation \cite{Pitler:2008tt}, the classification of implicit discourse relations has proved to be notoriously hard and it has remained one of the last missing pieces in an end-to-end discourse parser \cite{conllst2015}. In the absence of explicit discourse connectives, implicit discourse relations have to be inferred from their two arguments. Previous approaches on inferring implicit discourse relations have typically relied on features extracted from their two arguments. These features include word pairs that are the Cartesian products of the word tokens in the two arguments as well as features manually crafted from various lexicons such as verb classes and sentiment lexicons \cite{Pitler:2009th,rutherford2014brown}. These lexicons are used mainly to offset the data sparsity problem created by pairs of word tokens used directly as features. 
 
 Neural network models are an attractive alternative for this task for at least two reasons. First, they can model the argument of an implicit discourse relation as dense vectors and suffer less from the data sparsity problem that is typical of the traditional feature engineering paradigm. Second, they should be easily extended to other languages as they do not require human-annotated lexicons. However, despite the many nice properties of neural network models, it is not clear how well they will fare with a small dataset, typicalley found in discourse annotation projects. Moreover, it is not straightforward to construct a single vector that properly represents the ``semantics'' of the arguments. As a result, neural network models that use dense vectors have been shown to have inferior performance against traditional systems that use manually crafted features, unless the dense vectors are combined with the hand-crafted surface features \cite{ji2015recursive}.
 
 In this work, we explore multiple neural architectures in an attempt to find the best distributed representation and neural network architecture suitable for this task in both English and Chinese. We do this by probing the different points on the spectrum of structurality from structureless bag-of-words models to sequential and tree-structured models. We use feedforward, sequential long short-term memory (LSTM), and tree-structured LSTM models to represent these three points on the spectrum. To the best of our knowledge, there is no prior study that investigates the contribution of the different architectures in neural discourse analysis. 
 
Our main contributions and findings from this work can be summarized as follows:
\begin{itemize}
	\item Our neural discourse model performs comparably with or even outperforms systems with surface features across different fine-grained discourse label sets.
	\item We investigate the contribution of the linguistic structures in neural discourse modeling and found that high-dimensional word vectors trained on a large corpus can compensate for the lack of structures in the model, given the small amount of annotated data. 
	\item We found that modeling the interaction across arguments via hidden layers is essential to improving the performance of an implicit discourse relation classifier.
	\item We present the first neural CDTB-style Chinese discourse parser, confirming that our current results and other previous findings conducted on English data also hold cross-linguistically.
\end{itemize}

\section{Related Work}
The prevailing approach for this task is to use surface features derived from various semantic lexicons \cite{Pitler:2009th}, reducing the number of parameters by mapping raw word tokens in the arguments of discourse relations to a limited number of entries in a semantic lexicon such as polarity and verb classes. 

Along the same vein, Brown cluster assignments have also been used as a general purpose lexicon that requires no human manual annotation \cite{rutherford2014brown}. However, these solutions still suffer from the data sparsity problem and almost always require extensive feature selection to work well \cite{Park:2012tk,lin2009recognizing,ji2015recursive}. The work we report here explores the use of the expressive power of distributed representations to overcome the data sparsity problem found in the traditional feature engineering paradigm. 

Neural network modeling has attracted much attention in the NLP community recently and has been explored to some extent in the context of this task. Recently, Braud and Denis \shortcite{braud2015embedding} tested various word vectors as features for implicit discourse relation classification and show that distributed features achieve the same level of accuracy as one-hot representations in some experimental settings. Ji et al. \shortcite{ji2015recursive,ji2016latent} advance the state of the art for this task using recursive and recurrent neural networks. In the work we report here, we systematically explore the use of different neural network architectures and show that when high-dimensional word vectors are used as input, a simple feed-forward architecture can outperform more sophisticated architectures such as sequential and tree-based LSTM networks, given the small amount of data.

Recurrent neural networks, especially LSTM networks, have changed the paradigm of deriving distributed features from a sentence \cite{hochreiter1997lstm}, but they have not been much explored in the realm of discourse parsing. LSTM models have been notably used to encode the meaning of source language sentence in neural machine translation \cite{cho2014neuralmt,devlin2014nnmt} and recently used to encode the meaning of an entire sentence to be used as features \cite{kiros2015skipthought}. Many neural architectures have been explored and evaluated, but there is no single technique that is decidedly better across all tasks. The LSTM-based models such as Kiros et al. \shortcite{kiros2015skipthought} perform well across tasks but do not outperform some other strong neural baselines. Ji et al. \shortcite{ji2016latent} uses a joint discourse language model to improve the performance on the coarse-grained label in the PDTB, but in our case, we would like to deduce how well LSTM fares in fine-grained implicit discourse relation classification. A joint discourse language model might not scale well to finer-grained label set, which is more practical for application.

\section{Model Architectures}

\begin{figure}[t]
\centering
\includegraphics[width=3.20in]{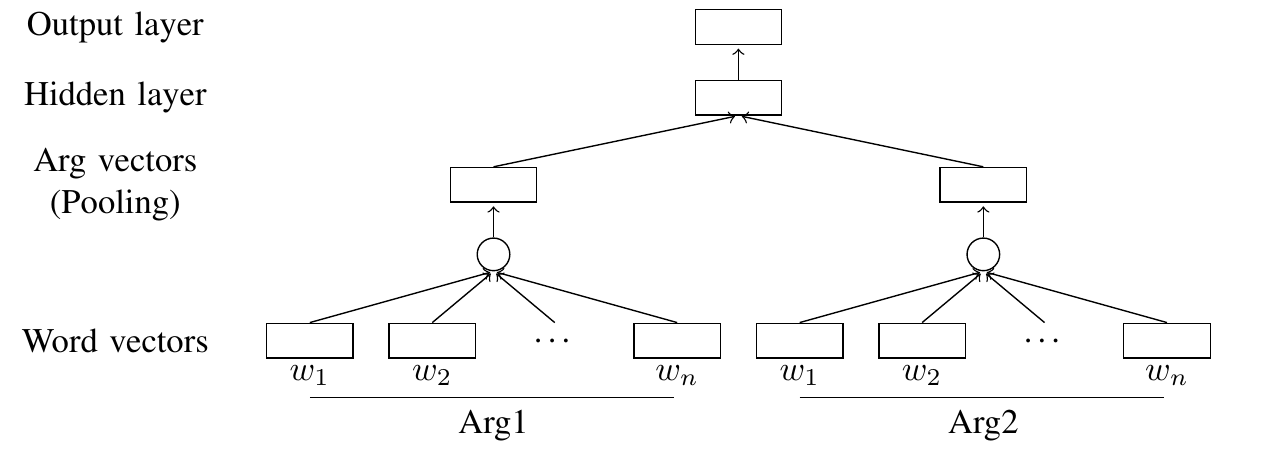}
\includegraphics[width=3.20in]{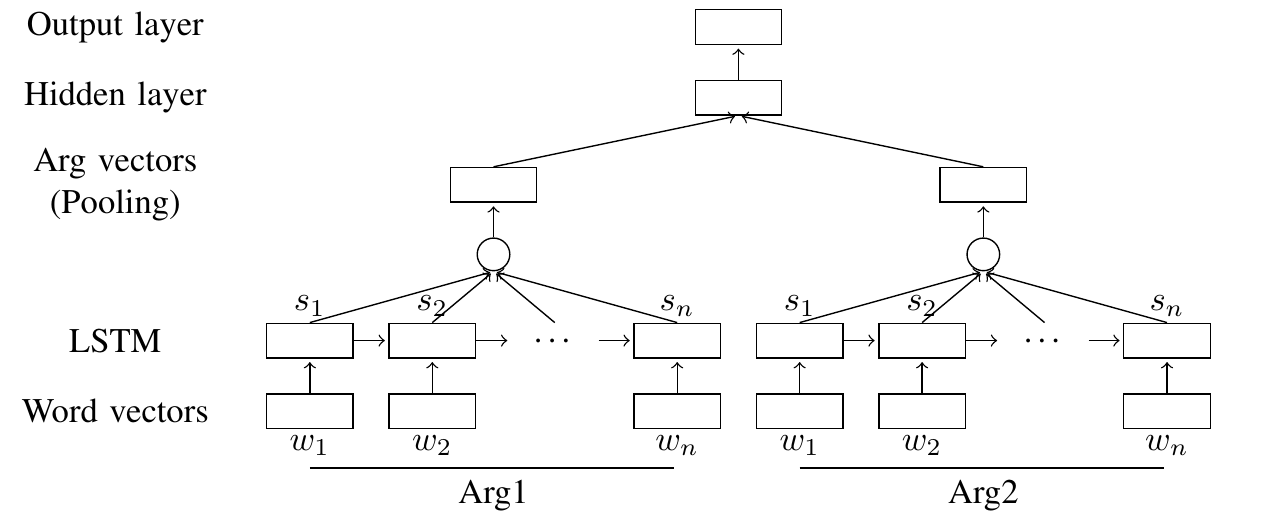}
\caption{(Left) Feedforward architecture. (Right) Sequential Long Short-Term Memory architecture.}
\label{model_architectures}
\end{figure}
Following previous work, we assume that the two arguments of an implicit discourse relation are given so that we can focus on predicting the senses of the implicit discourse relations. The input to our model is a pair of text segments called Arg1 and Arg2, and the label is one of the senses defined in the Penn Discourse Treebank as in the example below:  
	\begin{tabular}{lp{6.5cm}}
	\multicolumn{2}{l} {\textbf{Input:}}    \\
		Arg1 & Senator Pete Domenici calls this effort ``the first gift of democracy" \\
		Arg2 & The Poles might do better to view it as a \\
		     & Trojan Horse. \\
	\multicolumn{2}{l} {\textbf{Output:}}    \\
		Sense  & Comparison.Contrast\\
	\end{tabular}
In all architectures, each word in the argument is represented as a $k$-dimensional word vector trained on an unannotated data set. We use various model architectures to transform the semantics represented by the word vectors into distributed continuous-valued features. In the rest of the section, we explain the details of the neural network architectures that we design for the implicit discourse relations classification task. The models are summarized schematically in Figure \ref{model_architectures}.

\subsection{Bag-of-words Feedforward Model}
This model does not model the structure or word order of a sentence. The features are simply obtained through element-wise pooling functions. Pooling is one of the key techniques in neural network modeling of computer vision \cite{imagenet,cnn}. Max pooling is known to be very effective in vision, but it is unclear what pooling function works well when it comes to pooling word vectors. Summation pooling and mean pooling have been claimed to perform well at composing meaning of a short phrase from individual word vectors \cite{le2014paragraphvector,blacoe2012comparison,mikolov2013distributed,braud2015embedding}. The Arg1 vector $a^1$ and Arg2 vector $a^2$ are computed by applying element-wise pooling function $f$ on all of the $N_1$ word vectors in Arg1 $w^1_{1:N_1}$ and all of the $N_2$ word vectors in Arg2 $w^2_{1:N_2}$ respectively:
\begin{eqnarray*}
a^1_i = f(w^1_{1:N_1,i})\\
a^2_i = f(w^2_{1:N_2,i})	
\end{eqnarray*}

We consider three different pooling functions namely max, summation, and mean pooling functions: 
\begin{eqnarray*}
	f_{max}(w_{1:N},i) &=& \max_{j=1}^N w_{j,i} \\
	f_{sum}(w_{1:N},i) &=& \sum_{j=1}^N w_{j,i} \\
	f_{mean}(w_{1:N},i) &=& \sum_{j=1}^N w_{j,i} / N
\end{eqnarray*}

Inter-argument interaction is modeled directly by the hidden layers that take argument vectors as features. Discourse relations cannot be determined based on the two arguments individually. Instead, the sense of the relation can only be determined when the arguments in a discourse relation are analyzed jointly. The first hidden layer $h_1$ is the non-linear transformation of the weighted linear combination of the argument vectors:
$$h_1 = \tanh(W_1 \cdot a^1 + W_2 \cdot a^2 + b_{h_1})$$
where $W_1$ and $W_2$ are $d \times k$ weight matrices and $b_{h_1}$ is a $d$-dimensional bias vector. Further hidden layers $h_t$ and the output layer $o$ follow the standard feedforward neural network model. 
\begin{eqnarray*}
h_t = \tanh(W_{h_t} \cdot h_{t-1} + b_{h_t})\\
o = \mbox{softmax}(W_o \cdot h_T + b_o)
\end{eqnarray*}
where $W_{h_t}$ is a $d \times d$ weight matrix, $b_{h_t}$ is a $d$-dimensional bias vector, and $T$ is the number of hidden layers in the network. 

\subsection{Sequential Long Short-Term Memory (LSTM)}
A sequential Long Short-Term Memory Recurrent Neural Network (LSTM-RNN) models the semantics of a sequence of words through the use of hidden state vectors. Therefore, the word ordering does affect the resulting hidden state vectors, unlike the bag-of-word model. For each word vector at word position $t$, we compute the corresponding hidden state vector $s_t$ and the memory cell vector $c_t$ from the previous step. 
\begin{eqnarray*}
	i_t &=& \mbox{sigmoid}(W_i \cdot w_t + U_i \cdot s_{t-1} + b_i)\\ 
	f_t &=& \mbox{sigmoid}(W_f \cdot w_t + U_f \cdot s_{t-1} + b_f)\\
	o_t &=& \mbox{sigmoid}(W_o \cdot w_t + U_o \cdot s_{t-1} + b_o)\\
	c'_t &=& \tanh(W_c \cdot w_t + U_c \cdot s_{t-1} + b_c)\\
	c_t &=& c'_t * i_t + c_{t-1} * f_t \\
	s_t &=& c_t * o_t
\end{eqnarray*}
where $*$ is elementwise multiplication. The argument vectors are the results of applying a pooling function over the hidden state vectors. 
\begin{eqnarray*}
a^1_i = f(s^1_{1:N_1,i})\\
a^2_i = f(s^2_{1:N_2,i})	
\end{eqnarray*}
In addition to the three pooling functions that we describe in the previous subsection, we also consider using only the last hidden state vector, which should theoretically be able to encode the semantics of the entire word sequence. 
$$f_{last}(s_{1:N,i}) = s_{N,i}$$
Inter-argument interaction and the output layer are modeled in the same fashion as the bag-of-words model once the argument vector is computed. 

\subsection{Tree LSTM}
The principle of compositionality leads us to believe that the semantics of the argument vector should be determined by the syntactic structures and the meanings of the constituents. For a fair comparison with the sequential model, we apply the same formulation of LSTM on the binarized constituent parse tree. The hidden state vector now corresponds to a constituent in the tree. These hidden state vectors are then used in the same fashion as the sequential LSTM. The mathematical formulation is the same as Tai et al. \shortcite{tai2015tlstm}.

This model is similar to the recursive neural networks proposed by Ji and Eisenstein (2015). Our model differs from their model in several ways. We use the LSTM networks instead of the ``vanilla" RNN formula and expect better results due to less complication with vanishing and exploding gradients during training. Furthermore, our purpose is to compare the influence of the model structures. Therefore, we must use LSTM cells in both sequential and tree LSTM models for a fair and meaningful comparison. The more in-depth comparison of our work and recursive neural network model by Ji and Eisenstein (2015) is provided in the discussion section. 

\section{Corpora and Implementation}
\begin{table}
\footnotesize
\centering
	\begin{tabular}{lccc}
	\hline \hline
Sense & Train & Dev & Test \\
\hline
Comparison.Concession & 192 & 5 & 5 \\
Comparison.Contrast & 1612 & 82 & 127 \\
Contingency.Cause & 3376 & 120 & 197 \\
Contingency.Pragmatic cause & 56 & 2 & 5 \\
Expansion.Alternative & 153 & 2 & 15 \\
Expansion.Conjunction & 2890 & 115 & 116 \\
Expansion.Instantiation & 1132 & 47 & 69 \\
Expansion.List & 337 & 5 & 25 \\
Expansion.Restatement & 2486 & 101 & 190 \\
Temporal.Asynchronous & 543 & 28 & 12 \\
Temporal.Synchrony & 153 & 8 & 5 \\
\hline 
Total & 12930 & 515 & 766 \\
\hline \hline
	\end{tabular}
	\label{label_dist}
	\caption{The distribution of the level 2 sense labels in the Penn Discourse Treebank. The instances annotated with two labels are not double-counted (only first label is counted here), and partial labels are excluded.}
\end{table}

\noindent \textbf{The Penn Discourse Treebank (PDTB)}  We use the PDTB due to its theoretical simplicity in discourse analysis and its reasonably large size. The annotation is done as another layer on the Penn Treebank on Wall Street Journal sections. Each relation consists of two spans of text that are minimally required to infer the relation, and the sense is organized hierarchically. The classification problem can be formulated in various ways based on the hierarchy. Previous work in this task has been done over three schemes of evaluation: top-level 4-way classification \cite{Pitler:2009th}, second-level 11-way classification \cite{lin2009recognizing,ji2015recursive}, and modified second-level classification introduced in the CoNLL 2015 Shared Task \cite{conllst2015}. We focus on the second-level 11-way classification because the labels are fine-grained enough to be useful for downstream tasks and also because the strongest neural network systems are tuned to this formulation. If an instance is annotated with two labels ($\sim$3\% of the data), we only use the first label. Partial labels, which constitute $\sim$2\% of the data, are excluded. Table~\ref{label_dist} shows the distribution of labels in the training set (sections 2-21), development set (section 22), and test set (section 23).

\noindent \textbf{Training} Weight initialization is uniform random, following the formula recommended by Bengio \shortcite{bengio2012practical}. The cost function is the standard cross-entropy loss function, as the hinge loss function (large-margin framework) yields consistently inferior results. We use Adagrad as the optimization algorithm of choice. The learning rates are tuned over a grid search. We monitor the accuracy on the development set to determine convergence and prevent overfitting. L2 regularization and/or dropout do not make a big impact on performance in our case, so we do not use them in the final results.  

\noindent \textbf{Implementation} All of the models are implemented in Theano \cite{theano2010,theano2012}. The gradient computation is done with symbolic differentiation, a functionality provided by Theano. Feedforward models and sequential LSTM models are trained on CPUs on Intel Xeon X5690 3.47GHz, using only a single core per model. A tree LSTM model is trained on a GPU on Intel Xeon CPU E5-2660. All models converge within hours. 

\section{Experiment on the Second-level Sense in the PDTB}
We want to test the effectiveness of the inter-argument interaction and the three models described above on the fine-grained discourse relations in English. The data split and the label set are exactly the same as previous works that use this label set \cite{lin2009recognizing,ji2015recursive}.

\noindent \textbf{Preprocessing} All tokenization is taken from the gold standard tokenization in the PTB \cite{marcus1993building}. We use the Berkeley parser to parse all of the data \cite{berkeleyparser}. We test the effects of word vector sizes. 50-dimensional and 100-dimensional word vectors are trained on the training sections of WSJ data, which is the same text as the PDTB annotation. Although this seems like too little data, 50-dimensional WSJ-trained word vectors have previously been shown to be the most effective in this task \cite{ji2015recursive}. Additionally, we also test the off-the-shelf word vectors trained on billions of tokens from Google News data freely available with the \texttt{word2vec} tool. All word vectors are trained on the Skip-gram architecture \cite{mikolov2013distributed,mikolov2013efficient}. Other models such as GloVe and continuous bag-of-words seem to yield broadly similar results \cite{pennington2014glove}. We keep the word vectors fixed, instead of fine-tuning during training.

\begin{table}[t]
\centering
\begin{tabular}{lc}
  \hline \hline 
Model & Accuracy \\ 
  \hline
   \noalign{\vskip 2mm}  
  \textit{PDTB Second-level senses} \\
  Most frequent tag baseline & 25.71 \\
  Our best tree LSTM & 34.07 \\ 
  Ji \& Eisenstein, (2015) & 36.98 \\ 
  Our best sequential LSTM variant & 38.38 \\ 
  Our best feedforward variant & 39.56 \\ 
  Lin et al., (2009) & 40.20 \\ 
   \hline \hline
\end{tabular}
\caption{Performance comparison across different models for second-level senses.}
\label{best_results}
\end{table}

\begin{figure}[t]
	\includegraphics{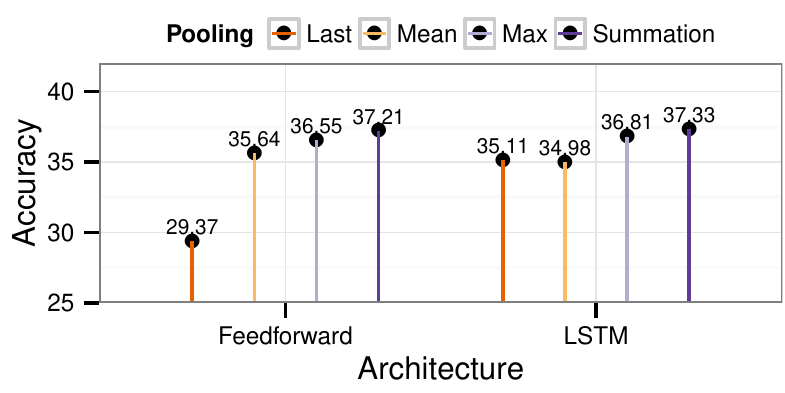}
	\caption{Summation pooling gives the best performance in general. The results are shown for the systems using 100-dimensional word vectors and one hidden layer.}
	\label{pooling}
\end{figure}

\begin{figure}[t]
	\includegraphics{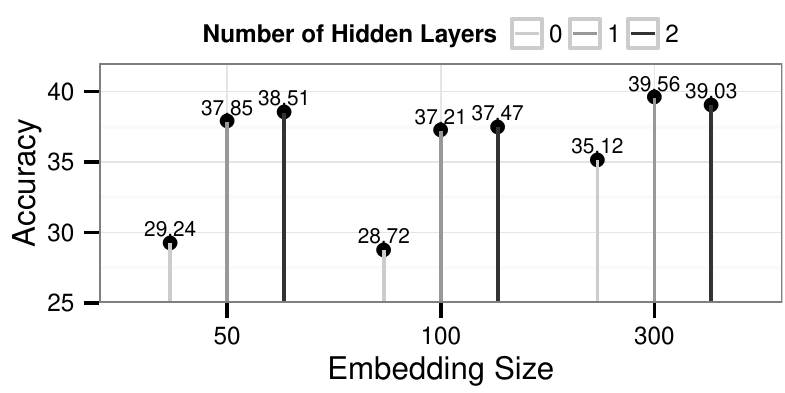}
    \caption{Inter-argument interaction can be modeled effectively with hidden layers. The results are shown for the feedforward models with summation pooling, but this effect can be observed robustly in all architectures we consider.}
	\label{hidden_layers}
\end{figure}

\begin{figure}[t]
	\includegraphics{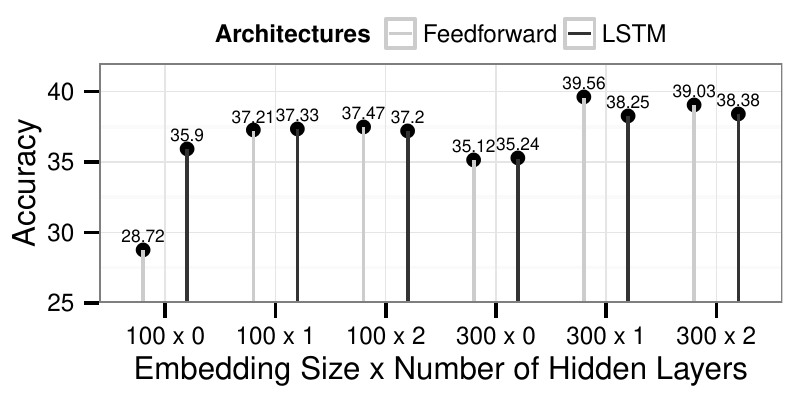}
	\caption{Comparison between feedforward and sequential LSTM when using summation pooling function.}
	\label{structures}
\end{figure}

\begin{table*}[t]
\footnotesize
\centering
\begin{tabular}{ll|llll|llll|llll|}
            &       & \multicolumn{4}{c}{No hidden layer} & \multicolumn{4}{c}{1 hidden layer} & \multicolumn{4}{c}{2 hidden layers} \\
Architecture & $k$   & max     & mean   & sum    & last   & max     & mean   & sum    & last   & max     & mean    & sum    & last   \\
            \hline 
Feedforward & 50    & 31.85   & 31.98  & 29.24  & -  & 33.28   & 34.98  & 37.85  & -   & 34.85   & 35.5    & 38.51  & -  \\
LSTM        & 50    & 31.85   & 32.11  & 34.46  & 31.85  & 34.07   & 33.15  & 36.16  & 34.34  & 36.16   & 35.11   & 37.2   & 35.24  \\
Tree LSTM   & 50    & 28.59   & 28.32  & 30.93  & 28.72  & 29.89   & 30.15  & 32.5   & 31.59  & 32.11   & 31.2    & 32.5   & 29.63  \\
Feedforward & 100   & 33.29   & 32.77  & 28.72  & -  & 36.55   & 35.64  & 37.21  & -  & 36.55   & 36.29   & 37.47  & -  \\
LSTM        & 100   & 30.54   & 33.81  & 35.9   & 33.02  & 36.81   & 34.98  & 37.33  & 35.11  & 37.46   & 36.68   & 37.2   & 35.77  \\
Tree LSTM   & 100   & 29.76   & 28.72  & 31.72  & 31.98  & 31.33   & 26.89  & 33.02  & 33.68  & 32.63   & 31.07   & 32.24  & 33.02  \\
Feedforward & 300   & 32.51   & 34.46  & 35.12  & -   & 35.77   & 38.25  & \textbf{39.56}  & -  & 35.25   & 38.51   & 39.03  & -  \\
LSTM        & 300   & 28.72   & 34.59  & 35.24  & 34.64  & 38.25   & 36.42  & 37.07  & 35.5   & \textbf{38.38}   & 37.72   & 37.2   & 36.29  \\
Tree LSTM   & 300   & 28.45   & 31.59  & 32.76  & 26.76  & 33.81   & 32.89  & 33.94  & 32.63  & 32.11   & 32.76   & \textbf{34.07}  & 32.50       
\end{tabular}

\caption{Compilation of all experimental configurations for 11-way classification on the PDTB test set. $k$ is the word vector size. Bold-faced numbers indicate the best performance for each architecture, which is also shown in Table \ref{best_results}. }
\label{all_experiments}
\end{table*}

\subsection{Results and discussion} 
The feedforward model performs best overall among all of the neural architectures we explore (Table \ref{best_results}). It outperforms the recursive neural network with bilinear output layer introduced by Ji and Eisenstein \shortcite{ji2015recursive} ($p<0.05$; bootstrap test) and performs comparably with the surface feature baseline \cite{lin2009recognizing}, which uses various lexical and syntactic features and extensive feature selection. Tree LSTM achieves inferior accuracy than our best feedforward model. The best configuration of the feedforward model uses 300-dimensional word vectors, one hidden layer, and the summation pooling function to derive argument feature vectors. The model behaves well during training and converges in less than an hour on a CPU. 

The sequential LSTM model outperforms the feedforward model when word vectors are not high-dimensional and not trained on a large corpus (Figure~\ref{structures}). Moving from 50 units to 100 units trained on the same dataset, we do not observe much of a difference in performance in both architectures, but the sequential LSTM model beats the feedforward model in both settings. This suggests that only 50 dimensions are needed for the WSJ corpus. However, the trend reverses when we move to 300-dimensional word vectors trained on a much larger corpus. These results suggest an interaction between the lexical information encoded by word vectors and the structural information encoded by the model itself. 

Hidden layers, especially the first one, make a substantial impact on performance. This effect is observed across all architectures (Figure~\ref{hidden_layers}). Strikingly, the improvement can be as high as 8\% absolute when used with the feedforward model with small word vectors. We tried up to four hidden layers and found that the additional hidden layers yield diminishing---if not negative---returns. These effects are not an artifact of the training process as we have tuned the models quite extensively, although it might be the case that we do not have sufficient data to fit those extra parameters. 

Summation pooling is effective for both feedforward and LSTM models (Figure~\ref{pooling}). The word vectors we use have been claimed to have some additive properties \cite{mikolov2013distributed}, so summation pooling in this experiment supports this claim. Max pooling is only effective for LSTM, probably because the values in the word vector encode the abstract features of each word relative to each other. It can be trivially shown that if all of the vectors are multiplied by -1, then the results from max pooling will be totally different, but the word similarities remain the same. The memory cells and the state vectors in the LSTM models transform the original word vectors to work well the max pooling operation, but the feedforward net cannot transform the word vectors to work well with max pooling as it is not allowed to change the word vectors themselves.

\subsection{Discussion}
Why does the feedforward model outperform the LSTM models? Sequential and tree LSTM models might work better if we are given larger amount of data. We observe that LSTM models outperform the feedforward model when word vectors are smaller, so it is unlikely that we train the LSTMs incorrectly. It is more likely that we do not have enough annotated data to train a more powerful model such as LSTM. In previous work, LSTMs are applied to tasks with a lot of labeled data compared to mere 12,930 instances that we have \cite{vinyals2015parsing,chiu2015cnnlstmner,irsoy2014deeprnn}. Another explanation comes from the fact that the contextual information encoded in the word vectors can compensate for the lack of structure in the model in this task. Word vectors are already trained to encode the words in their linguistic context especially information from word order.

Our discussion would not be complete without explaining our results in relation to the recursive neural network model proposed by Ji and Eisenstein \shortcite{ji2015recursive}. Why do sequential LSTM models outperform recursive neural networks or tree LSTM models? Although this first comes as a surprise to us, the results are consistent with recent works that use sequential LSTM to encode syntactic information. For example, Vinyals et al. \shortcite{vinyals2015parsing} use sequential LSTM to encode the features for syntactic parse output. Tree LSTM seems to show improvement when there is a need to model long-distance dependency in the data \cite{tai2015tlstm,li2015tlstm}. Furthermore, the benefits of tree LSTM are not readily apparent for a model that discards the syntactic categories in the intermediate nodes and makes no distinction between heads and their dependents, which are at the core of syntactic representations. 

Another point of contrast between our work and Ji and Eisenstein's \shortcite{ji2015recursive} is the modeling choice for inter-argument interaction. Our experimental results show that the hidden layers are an important contributor to the performance for all of our models. We choose linear inter-argument interaction instead of bilinear interaction, and this decision gives us at least two advantages. Linear interaction allows us to stack up hidden layers without the exponential growth in the number of parameters. Secondly, using linear interaction allows us to use high dimensional word vectors, which we found to be another important component for the performance. The recursive model by Ji and Eisenstein (2015) is limited to 50 units due to the bilinear layer. Our choice of linear inter-argument interaction and high-dimensional word vectors turns out to be crucial to building a competitive neural network model for classifying implicit discourse relations.

\section{Extending the results across label sets and languages}
Do our feedforward models perform well without surface features across different label sets and languages as well? We want to extend our results to another label set and language by evaluating our models on non-explicit discourse relation data used in English and Chinese CoNLL 2016 Shared Task. We will have more confidence in our model if it works well across label sets. It is also important that our model works cross-linguistically because other languages might not have resources such as semantic lexicons or parsers, required by some previously used features.

\subsection{English discourse relations}
We follow the experimental setting used in CoNLL 2015-2016 Shared Task as we want to compare our results against previous systems. This setting differs from the previous experiment in a few ways. Entity relations (EntRel) and alternative lexicalization relations (AltLex) are included in this setting. The label set is modified by the shared task organizers into 15 different senses including EntRel as another sense \cite{conllst2015}. We use the 300-dimensional word vector used in the previous experiment and tune the number of hidden layers and hidden units on the development set. The best results from last year's shared task are used as a strong baseline. It only uses surface features and also achieves the state-of-the-art performance under this label set \cite{wang2015conllst}. These features are similar to the ones used by Lin et al. (2009). 

\subsection{Chinese discourse relations}

We evaluate our model on the Chinese Discourse Treebank (CDTB) because its annotation is the most comparable to the PDTB \cite{zhou2015cdtb}. The sense set consists of 10 different senses, which are not organized in a hierarchy, unlike the PDTB. We use the version of the data provided to the CoNLL 2016 Shared Task participants. This version has 16,946 instances of discourse relations total in the combined training and development sets. The test set is not yet available at the time of submission, so the system is evaluated based on the average accuracy over 7-fold cross-validation on the combined set of training and development sets. 

There is no previously published baseline for Chinese. To establish baseline comparison, we use MaxEnt models loaded with the feature sets previously shown to be effective for English, namely dependency rule pairs, production rule pairs \cite{lin2009recognizing}, Brown cluster pairs \cite{rutherford2014brown}, and word pairs \cite{marcu2002unsupervised}. We use information gain criteria to select the best subset of each feature set, which is crucial in feature-based discourse parsing.

Chinese word vectors are induced through CBOW and Skipgram architecture in \texttt{word2vec} \cite{mikolov2013efficient} on Chinese Gigaword corpus \cite{chinese_gigaword} using default settings. The number of dimensions that we try are 50, 100, 150, 200, 250, and 300. We induce 1,000 and 3,000 Brown clusters on the Gigaword corpus. 

\begin{table}[t]
\centering
\begin{tabular}{lc}
\hline \hline
Model & Acc. \\ 
  \hline
  \noalign{\vskip 2mm}    
  \textit{CoNLL-ST 2015-2016 English} \\
  Most frequent tag baseline  & 21.36 \\
  Our best LSTM variant & 31.76 \\ 
  Wang and Lan (2015) - winning team & 34.45 \\
  Our best feedforward variant & \textbf{36.26} \\
  \noalign{\vskip 2mm}  
  \textit{CoNLL-ST 2016 Chinese} \\
  Most frequent tag baseline  & 77.14 \\
  ME + Production rules & 80.81 \\
  ME + Dependency rules & 82.34\\
  ME + Brown pairs (1000 clusters) & 82.36\\
  Out best LSTM variant & 82.48 \\
  ME + Brown pairs (3200 clusters) & 82.98 \\
  ME + Word pairs & 83.13 \\ 
  ME + All feature sets & 84.16 \\
  Our best feedforward variant & \textbf{85.45} \\ 
  \hline \hline
\end{tabular}
\caption{Our best feedforward variant significantly outperforms the systems with surface features ($p<0.05$). ME=Maximum Entropy model}
\label{conll_experiment}
\end{table}

\begin{figure}[t]
	\includegraphics{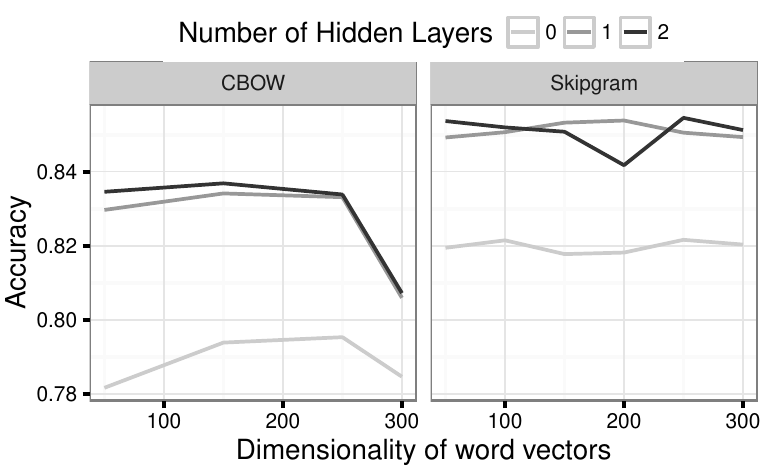}
	\caption{Comparing the accuracies across Chinese word vectors for feedforward model.}
	\label{chinese_word_vector}
\end{figure}

\subsection{Results} Table \ref{conll_experiment} shows the results for the models which are best tuned on the number of hidden units, hidden layers, and the types of word vectors. The feedforward variant of our model significantly outperforms the strong baselines in both English and Chinese ($p<0.05$ bootstrap test). This suggests that our approach is robust against different label sets, and our findings are valid across languages. Our Chinese model outperforms all of the feature sets known to work well in English despite using only word vectors. 

The choice of neural architecture used for inducing Chinese word vectors turns out to be crucial. Chinese word vectors from Skipgram model perform consistently better than the ones from CBOW model (Figure \ref{chinese_word_vector}). These two types of word vectors do not show much difference in the English tasks. 

\section{Conclusions and future work}
We report a series of experiments that systematically probe the effectiveness of various neural network architectures for the task of implicit discourse relation classification. Given the small amount of annotated data, we found that a feedforward variant of our model combined with hidden layers and high dimensional word vectors outperforms more complicated LSTM models. Our model performs better or competitively against models that use manually crafted surface features, and it is the first neural CDTB-style Chinese discourse parser. We will make our code and models publicly available. 

\bibliography{nlp_ref}
\bibliographystyle{acl2016}

\end{document}